\documentclass{article}
\usepackage[utf8]{inputenc}

\usepackage{graphicx}
\usepackage{comment}
\usepackage{amsmath}
\usepackage{amssymb}
\usepackage{color}
\usepackage{authblk}
\usepackage{float}
\usepackage{bm}
\usepackage{subcaption}
\usepackage{multirow}
\usepackage{wrapfig}
\usepackage{mathrsfs} 

\newcommand{\clip}[3]{\mathrm{clip}\left(#1,#2,#3\right)}
\newcommand{\round}[1]{\left\lceil#1\right\rfloor}
\newcommand{\vectorsym}[1]{\bm{#1}}

\newcommand{\expectation}[2]{\mathbb{E}_{#2}\left[#1\right]}
\newcommand{\bitset}{\mathrm{B}}
\newcommand{\threshset}{\mathrm{T}}
\newcommand{\mix}{{\tiny MP}}

\newcommand{\qb}{HMQ}

\title{HMQ: Hardware Friendly Mixed Precision Quantization Block for CNNs}

\author{\bf \small{Hai Victor Habi}}
\author{\bf \small{Roy H. Jennings}}
\author{\bf \small{Arnon Netzer}}
\affil{\large{Sony Semiconductor Israel}\\ \normalsize{\{hai.habi, roy.jennings, arnon.netzer\}@sony.com}}

\begin{document}
\maketitle
\thispagestyle{empty}

\begin{abstract}
Recent work in network quantization produced state-of-the-art results using mixed precision quantization.
An imperative requirement for many efficient edge device hardware implementations is that their quantizers are uniform and with power-of-two thresholds.
In this work, we introduce the Hardware Friendly Mixed Precision Quantization Block (HMQ) in order to meet this requirement.
The HMQ is a mixed precision quantization block that repurposes the Gumbel-Softmax estimator into a smooth estimator of a pair of quantization parameters, namely, bit-width and threshold.
HMQs use this to search over a finite space of quantization schemes.
Empirically, we apply HMQs to quantize classification models trained on CIFAR10 and ImageNet. 
For ImageNet, we quantize four different architectures and show that, in spite of the added restrictions to our quantization scheme, we achieve competitive and, in some cases, state-of-the-art results.
\end{abstract}

\section{Introduction}
{\let\thefootnote\relax\footnote{{The code of this work is available in https://github.com/sony-si/ai-research.}}}
In recent years, convolutional neural networks (CNNs) produced state-of-the-art results in many computer vision tasks including image classification \cite{gholami2018squeezenext,he2016deep,howard2017mobilenets,hu2018squeeze,sandler2018mobilenetv2,tan2019efficientnet}, object detection \cite{liu2016ssd,ren2015faster,tan2019efficientdet}, semantic segmentation \cite{long2015fully,ronneberger2015u}, etc. 
Deploying these models on embedded devices is a challenging task due to limitations on available memory, computational power and power consumption. 
Many works address these issues using different methods.
These include pruning \cite{han2015learning,yu2018nisp,zhang2018systematic}, efficient neural architecture design \cite{gholami2018squeezenext,howard2017mobilenets,jain2019trained,sandler2018mobilenetv2}, hardware and CNN co-design \cite{gholami2018squeezenext,howard2019searching,wu2019fbnet} and quantization \cite{cai2017deep,esser2019learned,han2015deep,jacob2018quantization,jain2019trained,zhang2018lq}. 

In this work, we focus on quantization, an approach in which the model is compressed by reducing the bit-widths of weights and activations.
Besides reduction in memory requirements, depending on the specific hardware, quantization usually also results in the reduction of both latency and power consumption.
The challenge of quantization is to reduce the model size without compromising its performance. 
For high compression rates, this is usually achieved by \emph{fine-tuning} a pre-trained model for quantization. 
In addition, recent work in quantization focused on making quantizers more \textit{hardware friendly} (amenable to deployment on embedded devices) by restricting quantization schemes to be: per-tensor, uniform, symmetric and with thresholds that are powers of two \cite{jain2019trained,Uhlich2020Mixed}.

Recently, \emph{mixed-precision} quantization was studied in \cite{dong2019hawq,Uhlich2020Mixed,wang2019haq,wu2018mixed}.
In these works, the bit-widths of weights and activations are not equal across the model and are learned during some optimization process.
In \cite{wang2019haq}, reinforcement learning is used, which requires the training of an agent that decides the bit-width of each layer.
In \cite{wu2018mixed}, neural architecture search is used, which implies duplication of nodes in the network and that the size of the model grows proportionally to the size of the search space of bit-widths.
Both of these methods limit the bit-width search space because of their computational cost.
In \cite{dong2019hawq}, the bit-widths are not searched during training, but rather, this method relies on the relationship between the layer's Hessian and its sensitivity to quantization.

An imperative requirement for many efficient edge device hardware implementations is that their quantizers are symmetric, uniform and with power-of-two thresholds (see \cite{jain2019trained}). 
This removes the cost of special handling of zero points and real value scale factors.
In this work, we introduce a novel quantization block we call the \textit{Hardware Friendly Mixed Precision Quantization Block} (\qb) that is designed to search over a finite set of quantization schemes that meet this requirement. 
\qb s utilize the Gumbel-Softmax estimator \cite{jang2016categorical} in order to optimize over a categorical distribution whose samples correspond to quantization scheme parameters.

We propose a method, based on \qb s, in which both the bit-width and the quantizer's threshold are searched simultaneously. 
We present state-of-the-art results on MobileNetV1, MobileNetV2 and ResNet-50 in most cases, in spite of the hardware friendly restriction applied to the quantization schemes.
Additionally, we present the first (that we know of) mixed precision quantization results of \mbox{EfficientNet-B0}.
In particular, our contributions are the following:
\begin{itemize}
	\item We introduce \qb, a novel, hardware friendly, mixed precision quantization block which enables a simple and efficient search for quantization parameters.
	\item We present an optimization method, based on \qb s, for mixed precision quantization in which we search simultaneously for both the bit-width and the threshold of each quantizer.
	\item We present competitive and, in most cases, state-of-the-art results using our method to quantize ResNet-50, EfficientNet-B0, MobileNetV1 and MobileNetV2 classification models on ImageNet.
\end{itemize}

\section{Related Work}\label{sec:related}
Quantization lies within an active area of research that tries to reduce memory requirements, power consumption and inference latencies of neural networks.
These works use techniques such as pruning, efficient network architectures and distillation (see e.g. \cite{chen2017learning,gholami2018squeezenext,han2015deep,he2019filter,he2017channel,howard2017mobilenets,liu2017learning,molchanov2016pruning,polino2018model,sandler2018mobilenetv2,tan2019efficientnet,zhang2018shufflenet}).
Quantization is a key method in this area of research which compresses and accelerates the model by reducing the number of bits used to represent model weights and activations.\\

\textbf{Quantization}. Quantization techniques can be roughly divided into two families: post-training quantization techniques and quantization-aware training techniques.
In post-training quantization techniques, a trained model is quantized without retraining the model (see e.g. \cite{banner2019post,cai2020zeroq}).
In quantization-aware training techniques, a model undergoes an optimization process during which the model is quantized.
A key challenge in this area of research, is to compress the model without significant degradation to its accuracy.
Post-training techniques suffer from a higher degradation to accuracy, especially for high compression rates.

Since the gradient of quantization functions is zero almost everywhere, most quantization-aware training techniques use the straight through estimator (STE) \cite{bengio2013estimating} for the estimation of the gradients of quantization functions.
These techniques mostly differ in their choice of quantizers, the quantizers' parametrization (thresholds, bit-widths, step size, etc.) and their training procedure. 
During training, the network weights are usually stored in full-precision and are quantized before they are used in feed-forward. 
The full-precision weights are then updated via back-propagation. 
Uniform quantizers are an important family of quantizers that have several benefits from a hardware point-of-view (see e.g. \cite{esser2019learned,jain2019trained,Uhlich2020Mixed}).
Non-uniform quantizers include clustering, logarithmic quantization and others (see e.g. \cite{baskin2018uniq,miyashita2016convolutional,zhang2018lq,zhou2017incremental}).\\

\textbf{Mixed precision}. Recent works on quantization produced state-of-the-art results using mixed precision quantization, that is, quantization in which the bit-widths are not constant across the model (weights and activations).
In \cite{wang2019haq}, reinforcement learning is used to determine bit-widths.
In \cite{dong2019hawq}, second order gradient information is used to determine bit-widths. 
More precisely, the bit-widths are selected by ordering the network layers using this information.
In \cite{Uhlich2020Mixed}, bit-widths are determined by learnable parameters whose gradients are estimated using STE. This work focuses on the choice of parametrization of the quantizers and shows that the threshold (dynamic range) and step size are preferable over parametrizations that use bit-widths explicitly.

In \cite{wu2018mixed}, a mixed precision quantization-aware training technique is proposed where the bit-widths search is converted into a network architecture search (based on \cite{liu2018darts}).
More precisely, in this solution, the search space of all possible quantization schemes is, in fact, a search for a sub-graph in a super-net.
The disadvantage of this approach, is that the size of the super net grows substantially with every optional quantization edge/path that is added to the super net.
In practice, this limits, the architecture search space.
Moreover, this work deals with bit-widths and thresholds as two separate problems where thresholds follow the solution in \cite{choi2018pact}.

\section{The HMQ Block}\label{sec:quant_block}
The \textit{Hardware Friendly Mixed Precision Quantization Block} (\qb) is a network block that learns, via standard SGD, a uniform and symmetric quantization scheme.
The scheme is parametrized by a pair $(t, b)$ of threshold $t$ and bit-width $b$.
During training, an \qb\ searches for $(t, b)$ over a finite space $\threshset\times \bitset\subseteq\mathbb{R}^+\times \mathbb{N}$.
In this work, we make \qb s ``hardware friendly" by also forcing their thresholds to be powers of two. 
We do this by restricting 
\begin{equation}\label{equ:threshold_set}
\threshset=\{2^M, 2^{M-1}, \dots, 2^{M-8}\}
\end{equation}
where $M\in\mathbb{Z}$ is an integer we configure per \qb\ (see Section \ref{sec:method}).\\

The step size $\Delta$ of a uniform quantization scheme is the (constant) gap between any two adjacent quantization points.
$\Delta$ is parametrized by $(t, b)$ differently for a signed quantizer, where $\Delta = \frac{2t}{2^{b}}$, and an unsigned one, where $\Delta =\frac{t}{2^{b}}$. 
Note that $\Delta$ ties the bit-width and threshold values into a single parameter but $\Delta$ is not uniquely defined by them. 
The definition of the quantizer that we use in this work is similar to the one in \cite{jain2019trained}.
The signed version $Q^{\rm{s}}$ of a quantizer of an \qb\ is defined as follows:
\begin{equation}\label{equ:signed_quantizer}
Q^{\rm{s}}(x, \Delta, t)=\clip{\Delta\cdot\round{\frac{\vectorsym{x}}{\Delta}}}{-(t-\Delta)}{t}
\end{equation}
where $\clip{x}{a}{b}=\min(\max(x,a), b)$ and $\round{x}$ is the rounding function.
Similarly, the unsigned version $Q^{\rm{us}}$ is defined as follows:
\begin{equation}\label{equ:unsigned_quantizer}
Q^{\rm{us}}(x,\Delta,t)=\clip{\Delta\cdot\round{\frac{\vectorsym{x}}{\Delta}}}{0}{t-\Delta}.
\end{equation}
In the rest of this section we assume that the quantizer $Q$ of an \qb\ is signed, but it applies to both signed and unsigned quantizers.\\

In order to search over a discrete set, the \qb\ represents each pair in $T\times B$ as a sample of a categorical random variable of the Gumbel-Softmax estimator (see \cite{jang2016categorical,maddison2016concrete}).
This enables the \qb\ to search for a pair of threshold and bit-width.
The Gumbel-Softmax is a continuous distribution on the simplex that approximates categorical samples.
In our case, we use this approximation as a joint discreet probability distribution of thresholds and bit-widths \mbox{$P_{\threshset,\bitset}(\threshset\!=\!t,\bitset\!=\!b|g_{t,b})$} on $\threshset\times \bitset$:
\begin{equation}\label{equ:gumbel_softmax_estimator}
P_{\threshset,\bitset}(\threshset=t,\bitset=b|g_{t,b}) = \frac{\exp(\frac{\log(\hat\pi_{t, b} )+g_{t, b}}{\tau})}
{\sum_{t'\in\threshset}\sum_{b'\in\bitset}\exp(\frac{\log(\hat\pi_{t', b'})+g_{t', b'}}{\tau})}
\end{equation}
where $\hat\pi$ is a matrix of class probabilities whose entries $\hat\pi_{t, b}$ correspond to pairs in $\threshset\times \bitset$,
$g_{t, b}$ are random i.i.d. variables drawn from Gumbel(0, 1) and $\tau>0$ is a softmax temperature value.
We define $\hat\pi=\mathrm{softmax}(\pi)$ where $\pi$ is a matrix of trainable parameters $\pi_{t, b}$. 
This guarantees that the matrix $\hat\pi$ forms a categorical distribution.
\\

The quantizers in Equations \ref{equ:signed_quantizer} and \ref{equ:unsigned_quantizer} are well defined for any two real numbers $\Delta>0$ and $t>0$.
During training, in feed forward, we sample $g_{t,b}$ and use these samples in the approximation $P_{\threshset,\bitset}$ of a categorical choice.
The \qb\ parametrizes its quantizer $Q(x, \hat{\Delta}, \hat{t})$ using an \textbf{expected step size} $\hat{\Delta}$ and an \textbf{expected threshold} $\hat{t}$ that are defined as follows:
\begin{equation}\label{equ:expected_step_size}
\hat{\Delta}=
\sum_{t\in\threshset}\sum_{b\in\bitset}P_{\threshset,\bitset}(\threshset=t,\bitset=b|g_{t,b})\cdot \Delta_{t,b},
\end{equation}
\begin{equation}\label{equ:expected_threshold}
\hat{t}=
\sum_{t\in\threshset}P_{\threshset}(\threshset=t)\cdot t
\end{equation}
where $P_{\threshset}(\threshset=t)=\sum_{b'\in\bitset}P_{\threshset,\bitset}(\threshset=t,\bitset=b'|g_{t,b'})$ is the marginal distribution of thresholds and $\Delta_{t, b} = \frac{2t}{2^{b}}$.

%
%
%
%
%
%
%
%

\begin{figure}
	\centering
	\includegraphics[scale=0.2]{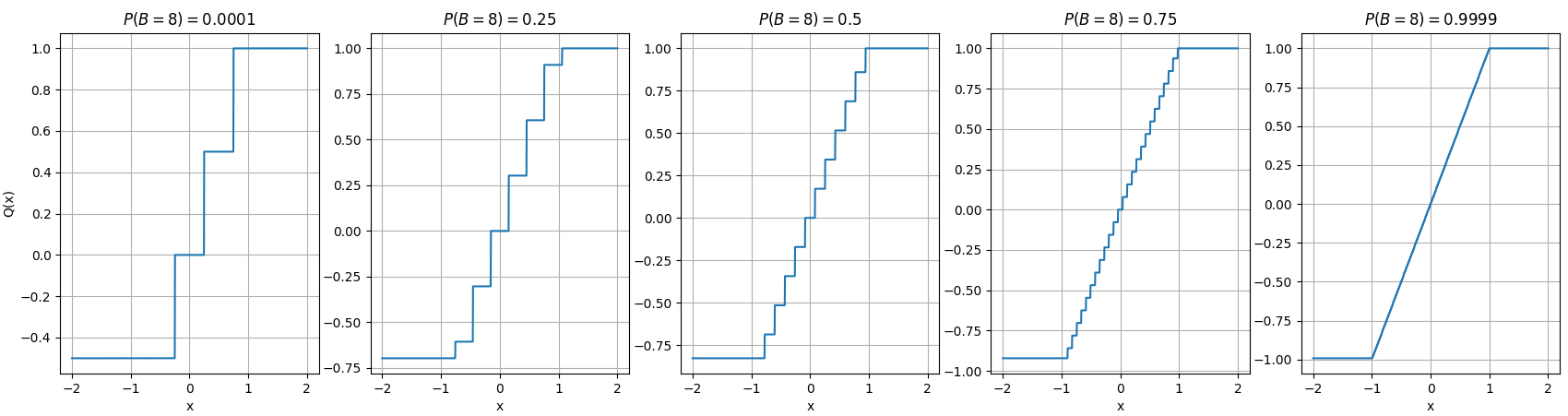}
	\caption{
		The quantization scheme of an \qb\ with $\threshset=\{1\}$ and $\bitset=\{2, 8\}$ for different approximations of the Gumbel-Softmax. 
		Transition from 2-bit quantization $P(\bitset=8)\approx 0$ (left) to 8-bit quantization $P(\bitset=8)\approx 1$ (right)
	}
	\label{fig:transition}
\end{figure}

In back-propagation, the gradients of rounding operations are estimated using the STE and the rest of the module, i.e. Equations \ref{equ:gumbel_softmax_estimator}, \ref{equ:expected_step_size} and \ref{equ:expected_threshold}, are differentiable.
This implies that the \qb\ smoothly updates the parameters $\pi_{t, b}$ which, in turn, smoothly updates the estimated bit-width and threshold values of the quantization scheme.
Figure \ref{fig:transition} shows examples of \qb\ quantization schemes during training.
During inference, the \qb's quantizer is parametrized by the pair $(t, b)$ that corresponds to the maximal parameter $\pi_{t, b}$.

Note that the temperature parameter $\tau$ of the Gumbel-Softmax estimator in Equation \ref{equ:gumbel_softmax_estimator} has a dual effect during training. 
As it approaches zero, in addition to approximating a categorical choice of a unique pair $(t, b)\in\threshset\times\bitset$, smaller values of $\tau$ also incur a larger variance of gradients which adds instability to the optimization process.
This problem is mitigated by annealing $\tau$ (see Section \ref{sec:method}).

\section{Optimization Process}\label{sec:method}
In this section, we present a fine-tuning optimization process that is applied to a full precision, \mbox{32-bit} floating point, pre-trained model after adding \qb s.
Throughout this work, we use the term model weights (or simply weights) to refer to all of the trainable model weights, \textit{not including} the \qb\ parameters.
We denote by $\Theta$, the set of weight tensors to be quantized; by $\mathcal{X}$, the set of activation tensors to be quantized and by $\Pi$, the set of \qb\ parameters.
Given a tensor $T$, we use the notation $|T|$ to denote the number of entries in $T$.\\

From a high level view, our optimization process consists of two phases.
In the first phase, we simultaneously train both the model weights and the \qb\ parameters.
We take different approaches for quantization of weights and activations.
These are described in Sections \ref{subsec:weight_compression} and \ref{subsec:activations_compression}.
We split the first phase into cycles with an equal number of epochs each.
In each cycle of the first phase, we reset the Gumbel-Softmax temperature $\tau$ in Equation \ref{equ:gumbel_softmax_estimator} and anneal it till the end of the cycle. 
In the second phase of the optimization process, we fine-tune only the model weights. 
During this phase, similarly to \qb s behaviour during inference, the quantizer of every \qb\ is parametrized by the pair $(t, b)$ that corresponds to the maximal parameter $\pi_{t, b}$ that was learnt in the first phase.


\subsection{Weight Compression}\label{subsec:weight_compression}
Let $\theta$ be an input tensor of weights to be quantized by some \qb.
We define the set of thresholds $\threshset$ in the search space $\threshset\times \bitset$ of the \qb\ by setting $M$ in Equation \ref{equ:threshold_set} to be $\min\{M:2^M\geq \max(\mathrm{abs}(\theta)), i\in\mathbb{Z}\}$.
The values in $\bitset$ are different per experiment (see Section \ref{se:experimental}).\\

Denote by $\Pi_w$ the subset of $\Pi$ containing all of the parameters of \qb s quantizing weights.
The expected weight compression rate, induced by the values of $\Pi_w$ is defined as follows:
\begin{equation}\label{equ:compression_rate}
\tilde{R}(\Pi_w)=\frac{32\sum_{\theta_i\in \Theta}|\theta_i|}{\sum_{\theta_i\in \Theta}\expectation{b_i}{}|\theta_i|}
\end{equation}
where $\theta_i$ is a tensor of weights and $\expectation{b_i}{}=\sum_{b\in\bitset}b\cdot P^{i}_{\bitset}(\bitset=b)$ is the expected bit-width of $\theta_i$, where $P^{i}_{B}$ is the bit-width marginal distribution in the Gumbel-Softmax estimation of the corresponding \qb.
In other words, assuming that all of the model weights are quantized by \qb s, the numerator is the memory requirement of the weights of the model before compression and the denominator is the expected memory requirement during training.\\

During the first phase of the optimization process, we optimize the model with respect to a target weight compression rate $R_w\in\mathbb{R^+}$, by minimizing (via standard SGD) the following loss function:
\begin{equation}\label{equ:loss}
J(\Theta,\Pi)=J_{task}(\Theta,\Pi)+\lambda \left(J_w(\Pi_w)\right)^{2}
\end{equation}
where $J_{task}(\Theta,\Pi)$ is the original, task specific loss, e.g. the standard cross entropy loss, $J_w(\Pi_w)$ is a loss with respect to the target compression rate $R_w$ and $\lambda$ is a hyper-parameter that control the trade-off between the two.
We define $J_{w}(\Pi_w)$ as follows:
\begin{equation}\label{equ:J_req}
J_{w}(\Pi_w)=\frac{\max(0, R_w-\tilde{R}(\Pi_w))}{R_w}.
\end{equation}
In practice, we gradually increase the target compression rate $R_w$ during the first few cycles in the first phase of our optimization process. 
This approach of gradual training of quantization is widely used, see e.g. \cite{baskin2018nice,dong2017learning,dong2019hawq,zhou2017incremental}.
In most cases, layers are gradually added to the training process whereas in our process we gradually decrease the bit-width across the whole model, albeit, with mixed precision.\\

By the definition of $J_{w}(\Pi_w)$, if the target weight compression rate is met during training, i.e. $\tilde{R}(\Pi_w) > R_w$, then the gradients of $J_{w}(\Pi_w)$ with respect to the parameters in $\Pi_w$ are zero and the task specific loss function determines the gradients alone. 
In our experiments, the actual compression obtained by using a specific target compression $R_w$ depends on the hyper-parameter $\lambda$ and the sensitivity of the architecture to quantization.


\subsection{Activations Compression}\label{subsec:activations_compression}
We define $\threshset$ in the search space $\threshset\times \bitset$ of an \qb\ that quantizes a tensor of activations similarly to \qb s quantizing weights.
We set $M\in\mathbb{Z}$ in \mbox{Equation \ref{equ:threshold_set}} to be minimum such that $2^M$ is greater or equal than the maximum absolute value of an activation of the pre-trained model over the entire training set.\\

The objective of activations compression is to fit any single activations tensor, after quantization, into a given size of memory $\overline{U}\in\mathbb{N}$ (number of bits).
This objective is inspired by the one in \cite{Uhlich2020Mixed} and is especially useful for DNNs in which the operators in the computational graph induce a path graph, i.e. the operators are executed sequentially. 
We define the target activations compression rate $R_a$ to be
\begin{equation}\label{equ:target_activation_compression}
R_a=\frac{32\cdot\max_{X_i\in\mathcal{X}}|X_i|}{\overline{U}}
\end{equation}
where $X_i$ are the activation tensors to be quantized.
Note that $\overline{U}$ implies the precise (maximum) number of bits $b(X)$ of every feature map $X\in\mathcal{X}$:
\begin{equation}\label{equ:act_bitwidth}
b(X)=\left\lfloor \frac{\overline{U}}{|X|}\right\rfloor.
\end{equation}
We assume that $b(X)\geq 1$ for every feature map $X\in\mathcal{X}$ (otherwise, the requirement cannot be met and $\overline{U}$ should be increased) and fix $\bitset=\{\min(b(X), 8)\}$ in the search space of the \qb\ that corresponds to $X$.
Note that this method can also be applied to models with a more complex computational graph, such as ResNet, by applying Equation \ref{equ:act_bitwidth} to blocks instead of single feature maps.
Note also, that by definition, the maximum bit-width of \textit{every activation} is $8$.
We can therefore assume that $R_a\geq 4$.\\

Here, the bit-widths of every feature map is determined by Equation \ref{equ:act_bitwidth}.
This is in contrast to the approach in \cite{Uhlich2020Mixed} (for activations compression) and our approach for weight compression in Section \ref{subsec:weight_compression}, where the choice of bit-widths is a result of an SGD minimization process.
This allows a more direct approach for the quantization of activations in which we gradually increase $R_a$, during the first few cycles in the first phase of the optimization process.
In this approach, while activation \qb s learn the thresholds, their bit-widths are implied by $R_a$.
This, in contrast to adding a target activations compression component to the loss, both guarantees that the target compression of activations is obtained and simplifies the loss function of the optimization process.

\section{Experimental Results}\label{se:experimental}
In this section, we present results using \qb s to quantize various classification models.
As proof of concept, we first quantize ResNet-18 \cite{he2016deep} trained on \mbox{CIFAR-10} \cite{krizhevsky2009learning}.
For the more challenging ImageNet \cite{deng2009imagenet} classification task, we present results quantizing ResNet-50 \cite{he2016deep}, EfficientNet-B0 \cite{tan2019efficientnet}, MobileNetV1 \cite{howard2017mobilenets} and MobileNetV2 \cite{sandler2018mobilenetv2}.
\\

In all of our experiments, we perform our fine-tuning process on a full precision, \mbox{32-bit} floating point, pre-trained model in which an \qb\ is added after every weight and every activation tensor per layer, \textit{including the first and last layers}, namely the input convolutional layer and the fully connected layer.  
The parameters $\pi_{t, b}$ of every \qb\ are initialized as a categorical distribution in which the parameter that corresponds to the pair of the maximum threshold with the maximum bit-width is initialized to $0.9$ and $0.1$ is uniformly distributed between the rest of the parameters.
The bit-width set $\bitset$ in the search space of \qb s is set differently for CIFAR-10 and ImageNet (see Sections \ref{subsec:cifar10} and \ref{subsec:imagenet}).\\

Note that in all of the experiments, in all of the weight \qb s, the maximal bit-width is 8 (similarly to activation \qb s). 
This implies that $\tilde{R}(\Pi_w)\geq 4$  throughout the fine-tuning process.
The optimizer that we use in all of our experiments is RAdam \cite{liu2019variance} with $\beta_1=0.9$ and $\beta_2=0.999$.
We use different learning rates for the model weights and the \qb\ parameters.
The data augmentation that we use during fine-tuning is the same as the one used to train the base models.
\\

The entire process is split into two phases, as described in Section \ref{sec:method}.
The first phase consists of $30$ epochs split into $6$ cycles of $5$ epochs each. 
In each cycle, the temperature $\tau$ in Equation \ref{equ:gumbel_softmax_estimator}, is reset and annealed till the end of the cycle.
We update the temperature every $N$ steps within a cycle, where $25\cdot N$ is the number of steps in a single epoch.
The annealing function that we use is similar to the one in \cite{jang2016categorical}:
\begin{equation}\label{eq:gumbel_temp}
\tau(i)=\max(e^{-ir},0.5)
\end{equation}
where $i$ is the training step (within the cycle) and $r=e^{-2}$.
The second phase, in which only weights are fine-tuned, consists of 20 epochs.\\

As mentioned in Section \ref{sec:method}, during the first phase, we gradually increase both the weight and activation target compression rates $R_w$ and $R_a$, respectively.
Both target compression rates are initialized to a minimum compression of $4$ (implying 8-bit quantization) and are increased, in equally sized steps, at the beginning of each cycle, during the first $4$ cycles.
\begin{figure}
	\centering
	\includegraphics[scale=0.3]{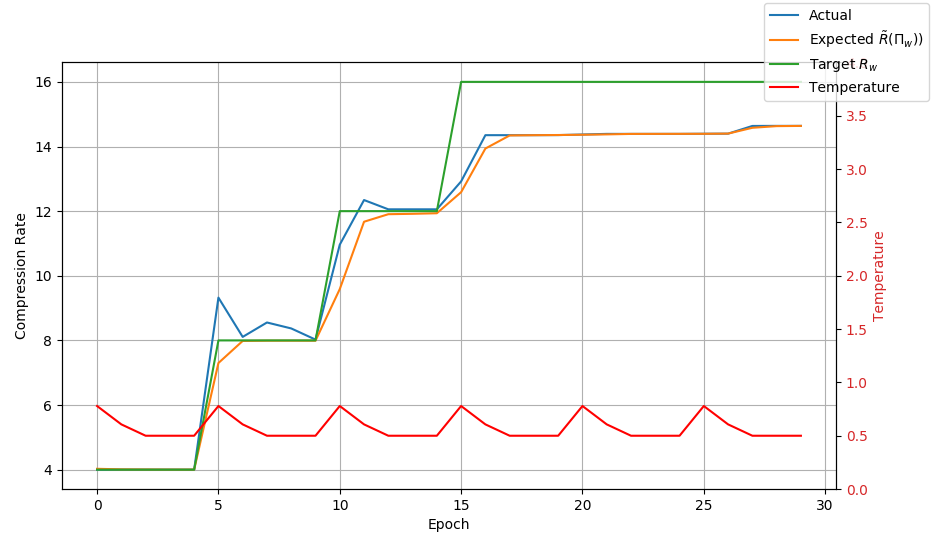}
	\caption{
		Expected and actual weight compression rates during fine-tuning of MobileNetV2 on ImageNet as the target compression rate and $\tau$ are updated
	}
	\label{fig:gradually_inc}
\end{figure}

Figure \ref{fig:gradually_inc} shows an example of the behaviour of the expected weight compression rate $\tilde{R}(\Pi_w)$ and the actual weight compression rate (implied by the quantization schemes corresponding to the maximum parameters $\pi_{t, b}$) during training, as the value of the target weight compression rate $R_w$ is increased and the temperature $\tau$ of the Gumbel-Softmax is annealed in every cycle.
Note how the difference between the expected and the actual compression rate values decreases with $\tau$, in every cycle (as to be expected by the Gumbel-Softmax estimator's behaviour).\\

We compare our results with those of other quantization methods based on top1 accuracy vs. compression metrics.
We use weight compression rate (WCR) to denote the ratio between the total size (number of bits) of the weights in the original model and the total size of the weights in the compressed model.
Activation compression rate (ACR) denotes the ratio between the size (number of bits) of the largest activation tensor in the original model and its size in the compressed model.
As explained in Section \ref{subsec:activations_compression}, our method guarantees that the size of every single activation tensor in the compressed model is bounded from above by a predetermined value $\overline{U}$.

\subsection{ResNet-18 on CIFAR-10}\label{subsec:cifar10}

As proof of concept, we use \qb s to quantize a ResNet-18 model that is trained on \mbox{CIFAR-10} with standard data-augmentation from \cite{he2016deep}. 
Our baseline model has top-1 accuracy of 92.45\%. 
We set $\bitset=\{1,2,3,4,5,6,7,8\}$ in the search space of \qb s quantizing weights.
For activations, $\bitset$ is set according to our method in Section \ref{subsec:activations_compression}. 
In all of the experiments in this section, we set $\lambda=32$ in the loss function in Equation \ref{equ:loss}.
The learning rate that we use for model weights is 1e-5. 
For \qb\ parameters the learning rate is 1e3.
The batch-size that we use is 256.\\


\begin{figure}[h]
	
	\begin{subfigure}{.48\textwidth}
		\centering
		\includegraphics[width=1.0\linewidth]{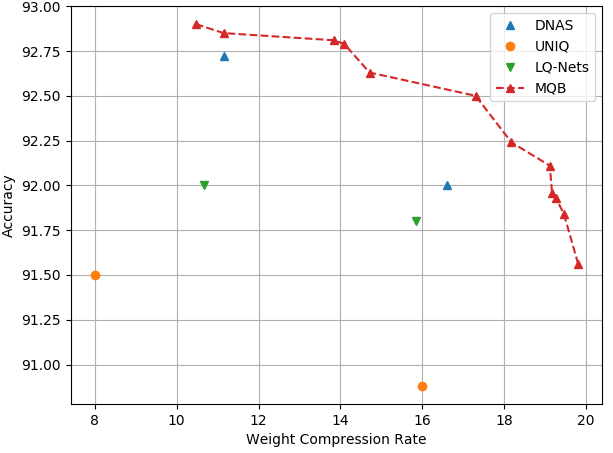}
		\caption{
			ACR$\approx$4
		}
		\label{fig:cifar10_parto_8bit}
	\end{subfigure}
	\hfill
	\begin{subfigure}{.48\textwidth}
		\centering
		\includegraphics[width=1.0\linewidth]{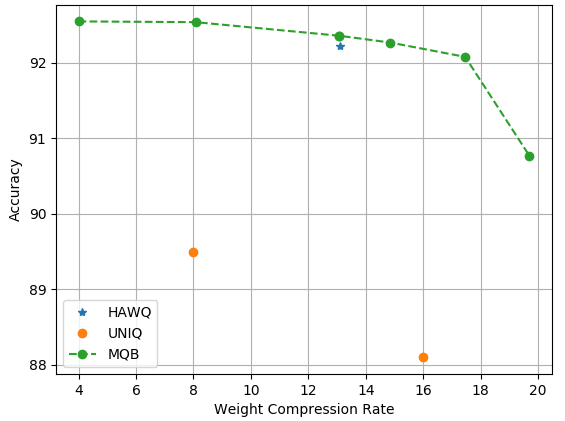}
		\caption{ACR$\approx$8}
		\label{fig:cifar10_parto_4bit}
	\end{subfigure}
	\caption{
		Pareto frontier of weight compression rate vs. top-1 accuracy of ResNet-18 on CIFAR-10 for two Activation Compression Rate (ACR) groups: $4$ \mbox{(Figure \ref{fig:cifar10_parto_8bit})} and 8 (Figure \ref{fig:cifar10_parto_4bit}) compared with different quantization methods
	}
	\label{fig:cifar10resnet-18} 
\end{figure}

Figure \ref{fig:cifar10resnet-18} presents the Pareto frontier of weight compression rate vs. top-1 accuracy for different quantization methods of ResNet-18 on CIFAR-10. 
In this figure, we show that our method is effective, in comparison to other methods, namely DNAS \cite{wu2018mixed}, UNIQ \cite{baskin2018uniq}, LQ-Nets \cite{zhang2018lq} and HAWQ \cite{dong2019hawq}, using different activation compression rates.

We explain our better results, compared to LQ-Nets and UNIQ, in-spite of the higher activation and weight compression rates, by the fact that \qb s take advantage of mixed precision quantization.
Compared to DNAS, our method has a much larger search space, since in their method, each quantization scheme is translated into a sub-graph in a super net. 
Moreover, \qb s tie the bit-width and threshold into a single parameter using Equation \ref{equ:expected_step_size}.
Comparing our method to HAWQ, HAWQ only uses the Hessian information whereas we perform an optimization over the bit-width.

\subsection{ImageNet}\label{subsec:imagenet}
In this section, we present results using \qb s to quantize several model architectures, namely MobileNetV1 \cite{howard2017mobilenets}, MobileNetV2 \cite{sandler2018mobilenetv2}, ResNet-50 \cite{he2016deep} and EfficientNet-B0 \cite{tan2019efficientnet} trained on the ImageNet \cite{deng2009imagenet} classification dataset.
In each of these cases, we use the same data augmentation as the one reported in the corresponding paper.
Our baseline models have the following top-1 accuracies: MobileNetV1 (70.6), MobileNetV2 (71.88\footnote{\label{foot:torchvision}Torchvision models (https://pytorch.org/docs/stable/torchvision/models.html)}), ResNet-50 (76.15\textsuperscript{\ref{foot:torchvision}}) and EfficientNet-B0 (76.8\footnote{https://github.com/tensorflow/tpu/tree/master/models/official/efficientnet}).
In all of the experiments in this section, we set $\bitset=\{2,3,4,5,6,7,8\}$ in the search space of \qb s quantizing weights.
For activations, $\bitset$ is set according to our method in Section \ref{subsec:activations_compression}. \\

As mentioned above, we use the RAdam optimizer in all of our experiments and we use different learning rates for the model weights and the \qb\ parameters.
For model weights, we use the following learning rates: MobileNetV1 (5e-6), MobileNetV2 (2.5e-6), ResNet-50 (2.5e-6) and EfficientNet-B0 (2.5e-6). 
For \qb\ parameters, the learning rate is equal to the learning rate of the weights multiplied by 1e3. 
The batch-sizes that we use are: MobileNetV1 (256), MobileNetV2 (128), ResNet-50 (64) and EfficientNet-B0 (128).


\subsubsection{Weight Quantization.}
In Table \ref{tab:comp_model_size}, we present our results using \qb s to quantize MobileNetV1, MobileNetV2 and ResNet-50.
In all of our experiments in this table, we set $R_a\!=\!4$ in Equation \ref{equ:target_activation_compression}, implying (single precision) 8-bit quantization of all of the activations. 
We split the comparison in this table into three compression rate groups: $\sim\!16$, $\sim\!10$ and $\sim\!8$ in rows 1--2, 3--4 and 5--6, respectively.
\begin{table}[h]
	\centering
	\caption{
		Weight Compression Rate (WCR) vs. top-1 accuracy (Acc) of MobileNetV1, MobileNetV2 and ResNet-50 on ImageNet.
		$R_w$ is the target weight compression rate in Equation \ref{equ:J_req} that was used for fine-tuning
	}
	\label{tab:comp_model_size}
\resizebox{\textwidth}{!}{%
	\begin{tabular}{lllllll}
		\hline
		\multirow{2}{*}{Method} & \multicolumn{2}{c}{MobileNetV1} & \multicolumn{2}{c}{MobileNetV2} & \multicolumn{2}{c}{ResNet-50} \\
		& WCR        & Acc         & WCR         & Acc        & WCR           & Acc           \\ \hline
		HAQ \cite{wang2019haq}  & 14.8       & 57.14       & 14.07       & 66.75      & 15.47         & 70.63         \\
		\qb\ (ours)                     & 14.15 {\tiny ($R_w\!=\!16$)}      & 68.36       & 14.4{\tiny ($R_w\!=\!16$)}        & 65.7       &  15.7  {\tiny ($R_w\!=\!16$)}           &  75             \\ \hline
		HAQ   & 10.22      & 67.66       & 9.68        & 70.9       & 10.41         & 75.30         \\
		\qb                      & 10.68 {\tiny ($R_w\!=\!11$)}      & 69.88       & 9.71 {\tiny ($R_w\!=\!10$)}        & 70.12      & 10.9 {\tiny ($R_w\!=\!11$)}         & 76.1          \\ \hline
		HAQ                    & 7.8        & 71.74       & 7.46        & 71.47      & 8             & 76.14         \\
		\qb                     & 7.6  {\tiny ($R_w\!=\!8$)}       & 70.912      & 7.7 {\tiny ($R_w\!=\!8$)}        & 71.4       & 9.01  {\tiny ($R_w\!=\!9$)}           & 76.3          \\ \hline
	\end{tabular}
}
\end{table}

Note that our method excels in very high compression rates.
Moreover, this is in spite of the fact that an \qb\ uses uniform quantization and its thresholds are limited to powers of two whereas HAQ uses k-means quantization.
We explain our better results by the fact that in HAQ, the bit-widths are the product of a reinforcement learning agent and the thresholds are determined by the statistics, opposed to \qb s, where they are the product of SGD optimization.

\subsubsection{Weight and Activation Quantization.}
In Table \ref{tab:comp_mix}, we compare mixed precision quantization methods in which both weights and activations are quantized. 
In all of the experiments in this table, the activation compression rate is equal to 8.
This means (with some variation between methods) that the smallest number of bits used to quantize activations is equal to 4.
This table shows that our method achieves on par results with other mixed precision methods, in spite of the restrictions on the quantization schemes of \qb s. 
We believe that this is due to the fact that, during training, there is no gradient mismatch for \qb\ parameters (see Equations \ref{equ:expected_step_size} and \ref{equ:expected_threshold}). 
In other words, \qb s allow smooth propagation of gradients. 
Additionally, \qb s tie each pair of bit-width and threshold in their search space with a single trainable parameter (opposed to determining the two separately).
\begin{table}[h]
	\centering
	\caption{
		Comparing Activation Compression Rate (ACR), Weight Compression Rate (WCR) and top-1 accuracy (Acc) of MobileNetV2 and ResNet-50 on ImageNet using different mixed precision quantization techniques.
		Under ACR:
		for HAWQ and HAWQ-V2, 8 means that the maximum compression obtained for a single activation tensor is 8.
		For DQ and \qb, 8 means that the compression of the largest activation tensor is 8
	}
	\label{tab:comp_aw}
	\begin{subtable}{0.49\textwidth}
		\centering
		\subcaption{MobileNetV2}
		\label{tab:comp_mbv2}
\resizebox{\textwidth}{!}{%
		\begin{tabular}{cccc}
			\hline
			Method     & ACR  & WCR   & Acc   \\ \hline
			DQ \cite{Uhlich2020Mixed} & 8.05 & 8.53  & 69.74 \\
			\qb {\tiny($R_w\!=\!8$)} (ours) & 8    & 8.05     & 70.9      \\ \hline
		\end{tabular}
}
	\end{subtable}
	\begin{subtable}{0.49\textwidth}
		\centering
		\subcaption{ResNet-50}
		\label{tab:comp_resnet-50}
\resizebox{\textwidth}{!}{%
		\begin{tabular}{cccc}
			\hline
			Method     & ACR  & WCR   & Acc   \\ \hline
			HAWQ \cite{wang2019haq}      & 8   & 12.28 & 75.3   \\ 
			HAWQ-V2 \cite{dong2019hawq2}  & 8   & 12.24 & 75.7   \\
			\qb {\tiny($R_w\!=\!13$)} (ours)   & 8   & 13.1  & 75.45  \\
			\hline
		\end{tabular}
}
	\end{subtable}
	\label{tab:comp_mix}
\end{table}


\subsubsection{EfficientNet.}
\begin{figure}\centering
	\includegraphics[scale=0.4]{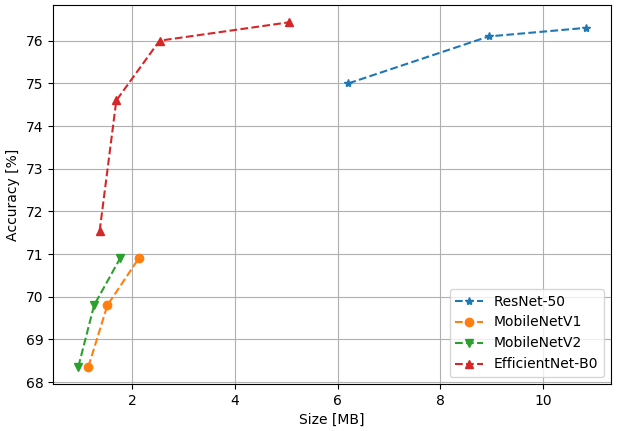}
	\caption{Pareto frontier of top-1 accuracy vs. model size of MobileNetV1, MobileNetV2, ResNet-50 and EfficientNet-B0 quantization by \qb}
	\label{fig:parto_imagenet}
\end{figure}

In Table \ref{tab:efficientnet_imagenet}, we present results quantizing EfficientNet-B0 using \qb s and in Figure \ref{fig:parto_imagenet}, we use the Pareto frontier of accuracy vs model size to summarize our results on all four of the models that were mentioned in this section. 

\begin{table}[h]
	\centering
	\caption{
		Weight Compression Rate (WCR) vs. top-1 accuracy (Acc) of EfficientNetB0 on ImageNet using \qb\ quantization.
		An Activation Compression Rate (ACR) of 4 means single precision 8-bit quantization of activation tensors.
		$R_w$ is the target weight compression rate that was used during fine-tuning
	}
	\label{tab:efficientnet_imagenet}
	\begin{tabular}{cccc}
		\hline
		ACR                & $R_w$ & WCR   & Acc   \\ \hline
		\multirow{4}{*}{4} & 4     & 4     & 76.4  \\
		& 8     & 8.05  & 76    \\
		& 12    & 11.97 & 74.6  \\
		& 16    & 14.87 & 71.54 \\ \hline
	\end{tabular}
\end{table}
\subsubsection{Additional Results.}
\begin{figure}[h]
	\centering
	\begin{subfigure}{.48\textwidth}
		\centering
		\includegraphics[width=1.0\linewidth]{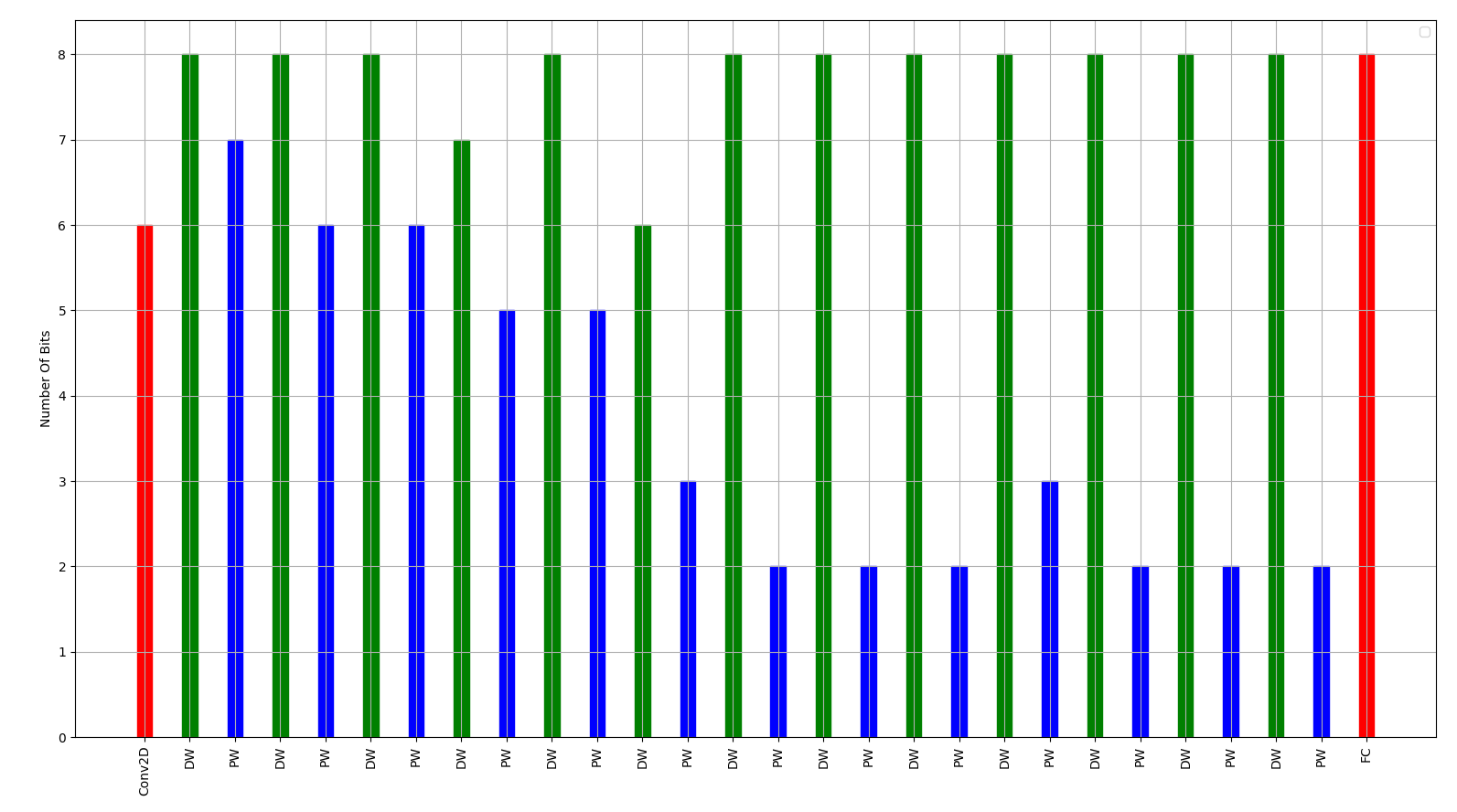}
		\caption{
			Weight bit-widths. 
			The red bars correspond to the first and last layers of the network. 
			The green bars correspond to depth-wise convolution layers and the blue bars correspond to point-wise convolution layers
		}
		\label{fig:mbv1_bit_width}
	\end{subfigure}
	\hfill
	\begin{subfigure}{.48\textwidth}
		\centering
		\includegraphics[width=1.0\linewidth]{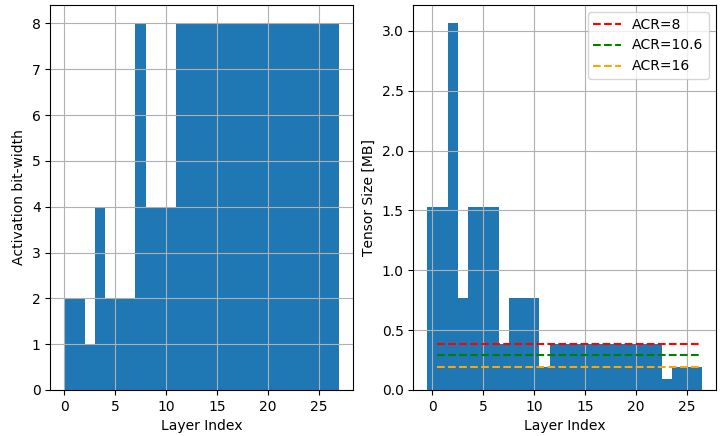}
		\caption{
			Activation bit-widths. 
			The right figure shows the sizes, per layer, of 32-bit activation tensors. 
			The dashed horizontal lines show the maximal tensor size implied by three target activation compression rates.
			The left figure shows the bit-widths, per layer (corresponding the right figure), at compression rate equal to 16
		}
		\label{fig:mbv1_bit_width_act}
	\end{subfigure}
	\caption{Example of the final bit-width of weights and activations in MobileNetV1 quantized by \qb} 
	\label{fig:mbv1_bit_width_example}
\end{figure}

In Figure \ref{fig:mbv1_bit_width_example}, we present an example of the final bit-widths of weights and activations in MobileNetV1 quantized by \qb.
This figure implies that point-wise convolutions are less sensitive to quantization, compared to their corresponding depth-wise convolutions.
Moreover, it seems that deeper layers are also less sensitive to quantization. 
Note that the bit-widths of activations in Figure \ref{fig:mbv1_bit_width_act} are not a result of fine-tuning but are pre-determined by the target activation compression, as described in Section \ref{subsec:activations_compression}.
In Table \ref{tab:imagenet_additional_results}, we present additional results using \qb s to quantize models trained on ImageNet.
This table extends the results in Table \ref{tab:comp_model_size}, here, both weights and activations are quantized using \qb s.
\begin{table}[H]
	\centering
	\caption{
		Weight Compression Rate (WCR) vs. top-1 accuracy (Acc) of MobileNet-V1, MobileNet-V2 and ResNet50 on ImageNet using \qb\ quantization with various target weight compression rates $R_w$ and a fixed Activation Compression Rate (ACR) of 8.
		MP means Mixed Precision
	}
	\label{tab:imagenet_additional_results}

		\begin{subtable}{0.49\textwidth}
		\subcaption{MobileNetV1}
		\centering
		\label{tab:comp_mbv1_ext}		
\resizebox{\textwidth}{!}{%
		\begin{tabular}{cccc}
			\hline
			$R_w$     & ACR & WCR    & Acc   \\ \hline
			16 & 8\mix  & 14.638\mix & 67.9  \\ \hline
			11 & 8\mix  & 10.709\mix & 69.3  \\ \hline
			\\
		\end{tabular}
}
	\end{subtable}
	\begin{subtable}{0.49\textwidth}
		\centering
		\subcaption{MobileNetV2}
		\label{tab:comp_mbv2_ext}
\resizebox{\textwidth}{!}{%
		\begin{tabular}{cccc}
			\hline
			$R_w$     & ACR  & WCR   & Acc   \\ \hline
			16 & 8\mix   & 14.8\mix     & 64.47     \\ \hline
			10& 8\mix    & 10\mix    & 69.9     \\ \hline
			\\
		\end{tabular}
}
	\end{subtable}

	\begin{subtable}{0.49\textwidth}
		\hfill
		\centering
		\subcaption{ResNet50}
		\label{tab:comp_resnet50_ext}
\resizebox{\textwidth}{!}{%
		\begin{tabular}{cccc}
			\hline
			$R_w$     & ACR  & WCR   & Acc   \\ \hline
			16 & 8\mix   & 15.45\mix     & 74.5     \\ \hline
			11& 8\mix    & 11.1\mix    & 75.73     \\ \hline
		\end{tabular}
}
	\end{subtable}

	\label{tab:comp_ext}
\end{table}

\section{Conclusions}\label{sec:conclusion}
In this work, we introduced the \qb, a novel quantization block that can be applied to weights and activations.
The \qb\ repurposes the Gumbel-Softmax estimator in order to smoothly search over a finite set of uniform and symmetric activation schemes. 
We presented a standard SGD fine-tuning process, based on \qb s, for mixed precision quantization that achieves state-of-the-art results in accuracy vs. compression for various networks.
Both the model weights and the quantization parameters are trained during this process.
This method can facilitate different hardware requirements, including memory, power and inference speed by configuring the \qb's search space and the loss function.
Empirically, we experimented with two image classification datasets: CIFAR-10 and ImageNet.
For ImageNet, we presented state-of-the-art results on MobileNetV1, MobileNetV2 and ResNet-50 in most cases.
Additionally, we presented the first (that we know of) quantization results of EfficientNet-B0.

\section*{Acknowledgments}\label{sec:ack}
We would like to thank Idit Diamant and Oranit Dror for many helpful discussions and suggestions.

\bibliographystyle{abbrv}
\bibliography{refs}

\begin{thebibliography}{10}

\bibitem{banner2019post}
R.~Banner, Y.~Nahshan, and D.~Soudry.
\newblock Post training 4-bit quantization of convolutional networks for
  rapid-deployment.
\newblock In {\em Advances in Neural Information Processing Systems}, pages
  7948--7956, 2019.

\bibitem{baskin2018nice}
C.~Baskin, N.~Liss, Y.~Chai, E.~Zheltonozhskii, E.~Schwartz, R.~Giryes,
  A.~Mendelson, and A.~M. Bronstein.
\newblock Nice: Noise injection and clamping estimation for neural network
  quantization.
\newblock {\em arXiv preprint arXiv:1810.00162}, 2018.

\bibitem{baskin2018uniq}
C.~Baskin, E.~Schwartz, E.~Zheltonozhskii, N.~Liss, R.~Giryes, A.~M. Bronstein,
  and A.~Mendelson.
\newblock Uniq: Uniform noise injection for non-uniform quantization of neural
  networks.
\newblock {\em arXiv preprint arXiv:1804.10969}, 2018.

\bibitem{bengio2013estimating}
Y.~Bengio, N.~L{\'e}onard, and A.~Courville.
\newblock Estimating or propagating gradients through stochastic neurons for
  conditional computation.
\newblock {\em arXiv preprint arXiv:1308.3432}, 2013.

\bibitem{cai2020zeroq}
Y.~Cai, Z.~Yao, Z.~Dong, A.~Gholami, M.~W. Mahoney, and K.~Keutzer.
\newblock Zeroq: A novel zero shot quantization framework.
\newblock In {\em Proceedings of the IEEE/CVF Conference on Computer Vision and
  Pattern Recognition}, pages 13169--13178, 2020.

\bibitem{cai2017deep}
Z.~Cai, X.~He, J.~Sun, and N.~Vasconcelos.
\newblock Deep learning with low precision by half-wave gaussian quantization.
\newblock In {\em Proceedings of the IEEE Conference on Computer Vision and
  Pattern Recognition}, pages 5918--5926, 2017.

\bibitem{chen2017learning}
G.~Chen, W.~Choi, X.~Yu, T.~Han, and M.~Chandraker.
\newblock Learning efficient object detection models with knowledge
  distillation.
\newblock In {\em Advances in Neural Information Processing Systems}, pages
  742--751, 2017.

\bibitem{choi2018pact}
J.~Choi, Z.~Wang, S.~Venkataramani, P.~I.-J. Chuang, V.~Srinivasan, and
  K.~Gopalakrishnan.
\newblock Pact: Parameterized clipping activation for quantized neural
  networks.
\newblock {\em arXiv preprint arXiv:1805.06085}, 2018.

\bibitem{deng2009imagenet}
J.~Deng, W.~Dong, R.~Socher, L.-J. Li, K.~Li, and L.~Fei-Fei.
\newblock Imagenet: A large-scale hierarchical image database.
\newblock In {\em 2009 IEEE conference on computer vision and pattern
  recognition}, pages 248--255. Ieee, 2009.

\bibitem{dong2017learning}
Y.~Dong, R.~Ni, J.~Li, Y.~Chen, J.~Zhu, and H.~Su.
\newblock Learning accurate low-bit deep neural networks with stochastic
  quantization.
\newblock {\em arXiv preprint arXiv:1708.01001}, 2017.

\bibitem{dong2019hawq2}
Z.~Dong, Z.~Yao, Y.~Cai, D.~Arfeen, A.~Gholami, M.~W. Mahoney, and K.~Keutzer.
\newblock Hawq-v2: Hessian aware trace-weighted quantization of neural
  networks.
\newblock {\em arXiv preprint arXiv:1911.03852}, 2019.

\bibitem{dong2019hawq}
Z.~Dong, Z.~Yao, A.~Gholami, M.~W. Mahoney, and K.~Keutzer.
\newblock Hawq: Hessian aware quantization of neural networks with
  mixed-precision.
\newblock In {\em Proceedings of the IEEE International Conference on Computer
  Vision}, pages 293--302, 2019.

\bibitem{esser2019learned}
S.~K. Esser, J.~L. McKinstry, D.~Bablani, R.~Appuswamy, and D.~S. Modha.
\newblock Learned step size quantization.
\newblock {\em arXiv preprint arXiv:1902.08153}, 2019.

\bibitem{gholami2018squeezenext}
A.~Gholami, K.~Kwon, B.~Wu, Z.~Tai, X.~Yue, P.~Jin, S.~Zhao, and K.~Keutzer.
\newblock Squeezenext: Hardware-aware neural network design.
\newblock In {\em Proceedings of the IEEE Conference on Computer Vision and
  Pattern Recognition Workshops}, pages 1638--1647, 2018.

\bibitem{han2015deep}
S.~Han, H.~Mao, and W.~J. Dally.
\newblock Deep compression: Compressing deep neural networks with pruning,
  trained quantization and huffman coding.
\newblock {\em arXiv preprint arXiv:1510.00149}, 2015.

\bibitem{han2015learning}
S.~Han, J.~Pool, J.~Tran, and W.~Dally.
\newblock Learning both weights and connections for efficient neural network.
\newblock In {\em Advances in neural information processing systems}, pages
  1135--1143, 2015.

\bibitem{he2016deep}
K.~He, X.~Zhang, S.~Ren, and J.~Sun.
\newblock Deep residual learning for image recognition.
\newblock In {\em Proceedings of the IEEE conference on computer vision and
  pattern recognition}, pages 770--778, 2016.

\bibitem{he2019filter}
Y.~He, P.~Liu, Z.~Wang, Z.~Hu, and Y.~Yang.
\newblock Filter pruning via geometric median for deep convolutional neural
  networks acceleration.
\newblock In {\em Proceedings of the IEEE Conference on Computer Vision and
  Pattern Recognition}, pages 4340--4349, 2019.

\bibitem{he2017channel}
Y.~He, X.~Zhang, and J.~Sun.
\newblock Channel pruning for accelerating very deep neural networks.
\newblock In {\em Proceedings of the IEEE International Conference on Computer
  Vision}, pages 1389--1397, 2017.

\bibitem{howard2019searching}
A.~Howard, M.~Sandler, G.~Chu, L.-C. Chen, B.~Chen, M.~Tan, W.~Wang, Y.~Zhu,
  R.~Pang, V.~Vasudevan, et~al.
\newblock Searching for mobilenetv3.
\newblock In {\em Proceedings of the IEEE International Conference on Computer
  Vision}, pages 1314--1324, 2019.

\bibitem{howard2017mobilenets}
A.~G. Howard, M.~Zhu, B.~Chen, D.~Kalenichenko, W.~Wang, T.~Weyand,
  M.~Andreetto, and H.~Adam.
\newblock Mobilenets: Efficient convolutional neural networks for mobile vision
  applications.
\newblock {\em arXiv preprint arXiv:1704.04861}, 2017.

\bibitem{hu2018squeeze}
J.~Hu, L.~Shen, and G.~Sun.
\newblock Squeeze-and-excitation networks.
\newblock In {\em Proceedings of the IEEE conference on computer vision and
  pattern recognition}, pages 7132--7141, 2018.

\bibitem{jacob2018quantization}
B.~Jacob, S.~Kligys, B.~Chen, M.~Zhu, M.~Tang, A.~Howard, H.~Adam, and
  D.~Kalenichenko.
\newblock Quantization and training of neural networks for efficient
  integer-arithmetic-only inference.
\newblock In {\em Proceedings of the IEEE Conference on Computer Vision and
  Pattern Recognition}, pages 2704--2713, 2018.

\bibitem{jain2019trained}
S.~R. Jain, A.~Gural, M.~Wu, and C.~Dick.
\newblock Trained quantization thresholds for accurate and efficient
  fixed-point inference of deep neural networks.
\newblock {\em arXiv preprint arXiv:1903.08066}, 2019.

\bibitem{jang2016categorical}
E.~Jang, S.~Gu, and B.~Poole.
\newblock Categorical reparametrization with gumble-softmax.
\newblock In {\em International Conference on Learning Representations (ICLR
  2017)}. OpenReview. net, 2017.

\bibitem{krizhevsky2009learning}
A.~Krizhevsky, G.~Hinton, et~al.
\newblock Learning multiple layers of features from tiny images.
\newblock Technical report, Citeseer, 2009.

\bibitem{liu2018darts}
H.~Liu, K.~Simonyan, and Y.~Yang.
\newblock {DARTS}: Differentiable architecture search.
\newblock In {\em International Conference on Learning Representations}, 2019.

\bibitem{liu2019variance}
L.~Liu, H.~Jiang, P.~He, W.~Chen, X.~Liu, J.~Gao, and J.~Han.
\newblock On the variance of the adaptive learning rate and beyond.
\newblock In {\em International Conference on Learning Representations}, 2020.

\bibitem{liu2016ssd}
W.~Liu, D.~Anguelov, D.~Erhan, C.~Szegedy, S.~Reed, C.-Y. Fu, and A.~C. Berg.
\newblock Ssd: Single shot multibox detector.
\newblock In {\em European conference on computer vision}, pages 21--37.
  Springer, 2016.

\bibitem{liu2017learning}
Z.~Liu, J.~Li, Z.~Shen, G.~Huang, S.~Yan, and C.~Zhang.
\newblock Learning efficient convolutional networks through network slimming.
\newblock In {\em Proceedings of the IEEE International Conference on Computer
  Vision}, pages 2736--2744, 2017.

\bibitem{long2015fully}
J.~Long, E.~Shelhamer, and T.~Darrell.
\newblock Fully convolutional networks for semantic segmentation.
\newblock In {\em Proceedings of the IEEE conference on computer vision and
  pattern recognition}, pages 3431--3440, 2015.

\bibitem{maddison2016concrete}
C.~J. Maddison, A.~Mnih, and Y.~W. Teh.
\newblock The concrete distribution: A continuous relaxation of discrete random
  variables.
\newblock In {\em International Conference on Learning Representations}, 2017.

\bibitem{miyashita2016convolutional}
D.~Miyashita, E.~H. Lee, and B.~Murmann.
\newblock Convolutional neural networks using logarithmic data representation.
\newblock {\em arXiv preprint arXiv:1603.01025}, 2016.

\bibitem{molchanov2016pruning}
P.~Molchanov, S.~Tyree, T.~Karras, T.~Aila, and J.~Kautz.
\newblock Pruning convolutional neural networks for resource efficient
  inference.
\newblock In {\em International Conference on Learning Representations}, 2017.

\bibitem{polino2018model}
A.~Polino, R.~Pascanu, and D.~Alistarh.
\newblock Model compression via distillation and quantization.
\newblock In {\em International Conference on Learning Representations}, 2018.

\bibitem{ren2015faster}
S.~Ren, K.~He, R.~Girshick, and J.~Sun.
\newblock Faster r-cnn: Towards real-time object detection with region proposal
  networks.
\newblock In {\em Advances in neural information processing systems}, pages
  91--99, 2015.

\bibitem{ronneberger2015u}
O.~Ronneberger, P.~Fischer, and T.~Brox.
\newblock U-net: Convolutional networks for biomedical image segmentation.
\newblock In {\em International Conference on Medical image computing and
  computer-assisted intervention}, pages 234--241. Springer, 2015.

\bibitem{sandler2018mobilenetv2}
M.~Sandler, A.~Howard, M.~Zhu, A.~Zhmoginov, and L.-C. Chen.
\newblock Mobilenetv2: Inverted residuals and linear bottlenecks.
\newblock In {\em Proceedings of the IEEE conference on computer vision and
  pattern recognition}, pages 4510--4520, 2018.

\bibitem{tan2019efficientnet}
M.~Tan and Q.~Le.
\newblock Efficientnet: Rethinking model scaling for convolutional neural
  networks.
\newblock In {\em International Conference on Machine Learning}, pages
  6105--6114, 2019.

\bibitem{tan2019efficientdet}
M.~Tan, R.~Pang, and Q.~V. Le.
\newblock Efficientdet: Scalable and efficient object detection.
\newblock {\em arXiv preprint arXiv:1911.09070}, 2019.

\bibitem{Uhlich2020Mixed}
S.~Uhlich, L.~Mauch, F.~Cardinaux, K.~Yoshiyama, J.~A. Garcia, S.~Tiedemann,
  T.~Kemp, and A.~Nakamura.
\newblock Mixed precision dnns: All you need is a good parametrization.
\newblock In {\em International Conference on Learning Representations}, 2020.

\bibitem{wang2019haq}
K.~Wang, Z.~Liu, Y.~Lin, J.~Lin, and S.~Han.
\newblock Haq: Hardware-aware automated quantization with mixed precision.
\newblock In {\em Proceedings of the IEEE Conference on Computer Vision and
  Pattern Recognition}, pages 8612--8620, 2019.

\bibitem{wu2019fbnet}
B.~Wu, X.~Dai, P.~Zhang, Y.~Wang, F.~Sun, Y.~Wu, Y.~Tian, P.~Vajda, Y.~Jia, and
  K.~Keutzer.
\newblock Fbnet: Hardware-aware efficient convnet design via differentiable
  neural architecture search.
\newblock In {\em Proceedings of the IEEE Conference on Computer Vision and
  Pattern Recognition}, pages 10734--10742, 2019.

\bibitem{wu2018mixed}
B.~Wu, Y.~Wang, P.~Zhang, Y.~Tian, P.~Vajda, and K.~Keutzer.
\newblock Mixed precision quantization of convnets via differentiable neural
  architecture search.
\newblock {\em arXiv preprint arXiv:1812.00090}, 2018.

\bibitem{yu2018nisp}
R.~Yu, A.~Li, C.-F. Chen, J.-H. Lai, V.~I. Morariu, X.~Han, M.~Gao, C.-Y. Lin,
  and L.~S. Davis.
\newblock Nisp: Pruning networks using neuron importance score propagation.
\newblock In {\em Proceedings of the IEEE Conference on Computer Vision and
  Pattern Recognition}, pages 9194--9203, 2018.

\bibitem{zhang2018lq}
D.~Zhang, J.~Yang, D.~Ye, and G.~Hua.
\newblock Lq-nets: Learned quantization for highly accurate and compact deep
  neural networks.
\newblock In {\em Proceedings of the European Conference on Computer Vision
  (ECCV)}, pages 365--382, 2018.

\bibitem{zhang2018systematic}
T.~Zhang, S.~Ye, K.~Zhang, J.~Tang, W.~Wen, M.~Fardad, and Y.~Wang.
\newblock A systematic dnn weight pruning framework using alternating direction
  method of multipliers.
\newblock In {\em Proceedings of the European Conference on Computer Vision
  (ECCV)}, pages 184--199, 2018.

\bibitem{zhang2018shufflenet}
X.~Zhang, X.~Zhou, M.~Lin, and J.~Sun.
\newblock Shufflenet: An extremely efficient convolutional neural network for
  mobile devices.
\newblock In {\em Proceedings of the IEEE conference on computer vision and
  pattern recognition}, pages 6848--6856, 2018.

\bibitem{zhou2017incremental}
A.~Zhou, A.~Yao, Y.~Guo, L.~Xu, and Y.~Chen.
\newblock Incremental network quantization: Towards lossless cnns with
  low-precision weights.
\newblock In {\em International Conference on Learning Representations}, 2017.

\end{thebibliography}
\end{document}